\documentclass[sigconf]{aamas}

\usepackage{balance} 

\makeatletter
\gdef\@copyrightpermission{
  \begin{minipage}{0.2\columnwidth}
   \href{https://creativecommons.org/licenses/by/4.0/}{\includegraphics[width=0.90\textwidth]{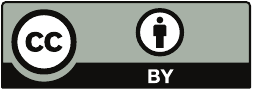}}
  \end{minipage}\hfill
  \begin{minipage}{0.8\columnwidth}
   \href{https://creativecommons.org/licenses/by/4.0/}{This work is licensed under a Creative Commons Attribution International 4.0 License.}
  \end{minipage}
  \vspace{5pt}
}
\makeatother

\setcopyright{ifaamas}
\acmConference[AAMAS '25]{Proc.\@ of the 24th International Conference
on Autonomous Agents and Multiagent Systems (AAMAS 2025)}{May 19 -- 23, 2025}
{Detroit, Michigan, USA}{Y.~Vorobeychik, S.~Das, A.~Nowé  (eds.)}
\copyrightyear{2025}
\acmYear{2025}
\acmDOI{}
\acmPrice{}
\acmISBN{}

\acmSubmissionID{<<1005>>}

\title[AAMAS-2025 Formatting Instructions]{HAVA: Hybrid Approach to Value Alignment through Reward Weighing for Reinforcement Learning}

\author{Kryspin Varys}
\affiliation{
  \institution{University of Southampton}
  \city{Southampton}
  \country{United Kingdom}}
\email{k.varys@soton.ac.uk}

\author{Federico Cerutti}
\affiliation{
  \institution{University of Brescia}
  \city{Brescia}
  \country{Italy}}
\email{federico.cerutti@unibs.it}

\author{Adam Sobey}
\affiliation{
  \institution{The Alan Turing Institute}
  \city{London}
  \country{United Kingdom}}
  \affiliation{
  \institution{University of Southampton}
  \city{Southampton}
  \country{United Kingdom}}
\email{asobey@turing.ac.uk}

\author{Timothy J. Norman}
\affiliation{
  \institution{University of Southampton}
  \city{Southampton}
  \country{United Kingdom}}
\email{t.j.norman@soton.ac.uk}

\begin{abstract}
Our society is governed by a set of norms which together bring about the values we cherish such as safety, fairness or trustworthiness. The goal of value alignment is to create agents that not only do their tasks but through their behaviours also promote these values. Many of the norms are written as laws or rules (legal / safety norms) but even more remain unwritten (social norms). Furthermore, the techniques used to represent these norms also differ. Safety / legal norms are often represented explicitly, for example, in some logical language while social norms are typically learned and remain hidden in the parameter space of a neural network. There is a lack of approaches in the literature that could combine these various norm representations into a single algorithm. We propose a novel method that integrates these norms into the reinforcement learning process. Our method monitors the agent's compliance with the given norms and summarizes it in a quantity we call the agent's reputation. This quantity is used to weigh the received rewards to motivate the agent to become value aligned. We carry out a two experiments including a continuous state space traffic problem to demonstrate the importance of the written and unwritten norms and show how our method can find the value aligned policies. Furthermore, we carry out ablations to demonstrate why it is better to combine these two groups of norms rather than using either separately.
\end{abstract}


\newcommand{\BibTeX}{\rm B\kern-.05em{\sc i\kern-.025em b}\kern-.08em\TeX}

\begin{document}


\pagestyle{fancy}
\fancyhead{}


\maketitle 


\section{Introduction}

Reinforcement learning (RL) is tackling increasingly complex tasks such as games \cite{Vinyals2019} and robotics \cite{10.1145/3528223.3530110}. and with these we also increasingly require the RL agents to respect the values of a society they operate in. For example, when AlphaStar \cite{Vinyals2019} learned to play StarCraft II, one of the conditions was a minimal amount of time, the agent had to spend ``looking at'' any location on the map before playing in a similar way a human player does. In other words, the aim shifted from trying to achieve the highest possible performance to achieving a high enough performance in a \textit{value aligned way}. In this case, \textit{the value} the agent had to promote was fair-play.

Machine ethics is concerned with creating machines whose behaviour is guided by a set of ethical principles \cite{anderson_anderson_2011}. Value alignment of artificial agents is an important task within machine ethics \cite{Balakrishnan_Bouneffouf_Mattei_Rossi_2019} whose goal is to design systems that achieve their intended objectives while remaining aligned with some values of the society they operate in \cite{10.1145/3375627.3375872}. These values could, for example, include safety, fairness, privacy or trustworthiness. Traditionally, a set of norms (rules) would be written down that promote the chosen values and passed to the agent. These approaches are known in the literature as top-down (rule-based) approaches \cite{Allen2005}. These norms would then be encoded within the reinforcement learning agent to motivate it to bring about the value while performing the task. This has been a popular method in safe reinforcement learning \cite{Alshiekh2018,10.1007/978-3-030-76384-8_15}. These methods work well for environments with relatively simple norms such as Pacman \cite{icaart22}. However, as the RL agents move from games to societies, from grid worlds to environments with continuous state and action spaces, the number of values and their associated norms rapidly increases and hard-coding them becomes prohibitively difficult.

To make matters worse, many of the societal norms are not clearly defined or written. That led researchers into techniques that aim to approximate norms or normative behaviours (preferences) from data. These approaches are known as bottom-up (data-driven) approaches \cite{Allen2005} and include fields such as inverse reinforcement learning \cite{ijcai2019p891, 9223344}, learning from human / AI feedback \cite{10.5555/3535850.3535966,casper2023open,lee2024rlaif} and teacher based methods \cite{Balakrishnan_Bouneffouf_Mattei_Rossi_2019}. A typical problem with these approaches is that we cannot verify that all important norms have been learned. While the rule-based approaches represent the norms explicitly, such that we can verify them, the data-driven approaches represent them implicitly often as parameters of a neural network. Furthermore, these approaches often fail to understand the norm severity; that some norms (e.g. safety / legal norms) could be considered mandatory while others (e.g. social norms) are tentative. In the cases when norm violation is unavoidable, we would prefer the agent to violate the less critical social norms rather than the safety / legal ones. However, for this an understanding of norm severity is necessary.

Some researchers believe that a combination of both rule-based and data-driven approaches to norm specification is the most promising avenue \cite{osti_10301363, Rossi_Mattei_2019}. Current approaches that allow to specify norms as well as learn them typically use the same technique for representing the norms \cite{Baert2023}. However, this might be a limitation \cite{Rossi_Mattei_2019} and approaches which allow combining norms represented in different ways are needed.

\begin{figure}[h]
  \centering
  \includegraphics[width=8.5cm]{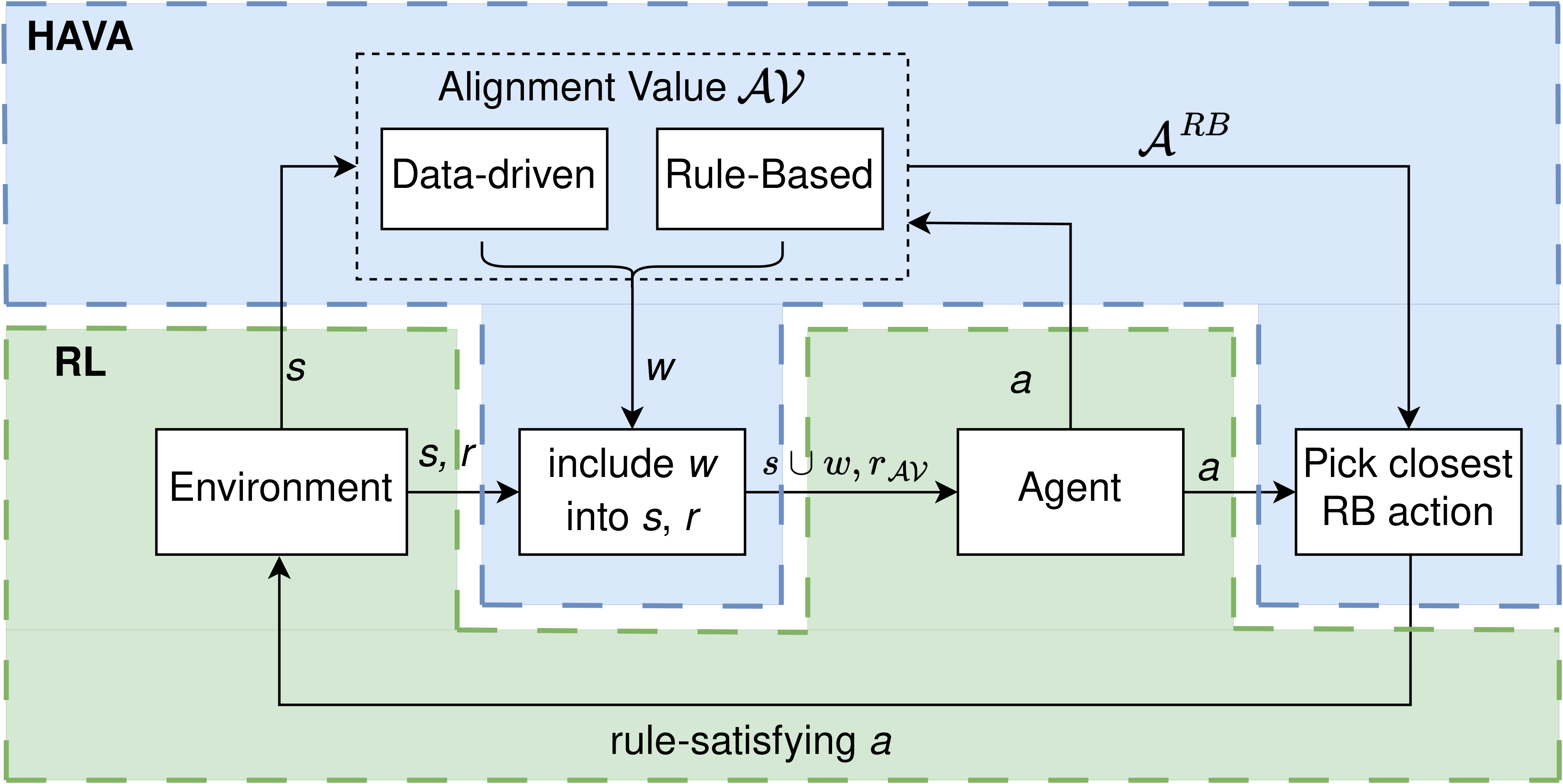}
  \caption{At each time-step Alignment Value \(\mathcal{AV}\) receives \(s_t, a_t\) from the agent and calculates the new agent's reputation \(w_{t+1}\) which then becomes part of the state \(s_{t+1}\). An agent's action is sent to the environment only if it is within the permitted rule-based actions. If not, another (closest) rule-based action is executed instead.}
  \label{fig:architecture}
\end{figure}

We propose a method called Hybrid Approach to Value Alignment (HAVA)\footnote{Demo available at \url{https://vimeo.com/1020279641?share=copy}. Code at \url{https://github.com/kvarys/HAVA}} (Figure \ref{fig:architecture}) that allows for both rule-based and data-driven norms. Furthermore, it allows these to be represented using different techniques. Our method divides the norms into two groups: 1) mandatory rule-based norms and 2) tentative data-driven norms. We expect the former to capture all of the safe / legal actions in some state of the environment, while the latter can serve as preferences the society has over these actions. If forced to choose, HAVA would always choose to violate the data-driven (social) norms and not the rule-based (safety / legal) norms. This can enable existing approaches to value alignment to be combined together under the HAVA framework to create new value aligned agents.

In HAVA a quantity is introduced, called the agent's \textit{reputation}, that keeps track of how much the agent's behaviour followed the passed norms. The reputation is then used to weigh the received task rewards forcing the agent to start following the norms. In this way HAVA side-steps the difficult task of manually engineering a reward that promotes aligned behaviours. While HAVA is governed by hyperparameters that the user has to set, we show these have an intuitive meaning that should aid in choosing their values.

We demonstrate the effectiveness of our approach on two experiments; one grid-world on which we intuitively describe the working of our method and one traffic scenario modelled in the SUMO simulator \cite{SUMO2018}. HAVA is compared to two ablations; a pure rule-based agent that lacks the data-driven norms and a pure data-driven agent that lacks the understanding that some norms are mandatory while others tentative. We show that the pure rule-based approach produces policies which are unsocial while the pure data-driven approach, if given a chance, might choose to violate safety / legal norms because to the data-driven agent all norms seem equally important.

Thus our main contribution is a novel method for value alignment in reinforcement learning that combines rule-based and data-driven norms, allows for these norms to be represented using different techniques and given the task reward function and hyperparameter values calculates a weight that motivates the agent to become value aligned.

Furthermore, through the ablations we show why social norms as well as understanding the norm severity are crucial for producing value aligned policies.

\section{Hybrid Approach to Value Alignment}

\subsection{Markov Decision Processes and Reinforcement Learning}
\label{sub:markov_decision_processes_and_reinforcement_learning}

Markov Decision Process (MDP) is a tuple \(\mathcal{M} = \left\langle \mathcal{S}, \mathcal{A}, T, R_{\text{task}}\right\rangle\) where \(\mathcal{S}\) is the set of states, \(\mathcal{A}\) is the set of actions, \(T: \mathcal{S} \times \mathcal{A} \rightarrow \mathcal{S}\) is the transition function and \(R_{\text{task}}: \mathcal{S} \times \mathcal{A} \rightarrow \mathbb{R}\) is the reward function specifying the agent's task in this MDP.

The goal of reinforcement learning (RL) is to find an agent's policy \(\pi: \mathcal{S} \rightarrow \mathcal{A}\) that maximizes the expected discounted return for each trajectory \(\sigma=\left(s_0,a_0,r_0,\dots \right)\):
\begin{align*}
  \max_{\pi \in \Pi} J(\pi) &:= \mathbb{E}_{\sigma\sim \pi} \left[\right. \textstyle\sum_{t=0}^{\infty}\gamma^t R_{\text{task}}(s_t, a_t)\left.\right]
\end{align*}
where \(\Pi\) is the policy space and \(\gamma \in [0, 1)\) is the discount factor.

\subsection{Values for Value Alignment in RL}
\label{sub:valuesystems}

\textbf{Definition 1:} The Alignment Value \(\mathcal{AV}=\left\langle RB, DD\right\rangle\) is a tuple where \(RB\) and \(DD\) are defined as follows:
\begin{enumerate}
  \item[] - \(RB: \mathcal{S} \rightarrow 2^{\mathcal{A}}\) is a function \(\mathcal{A}^{RB} = RB(s)\). We expect \(RB\) to be a rule-based system and to contain all the mandatory safety and legal norms. We then use \(RB\) to retrieve the permitted actions for any state \(s\in \mathcal{S}\). The RL agent is not allowed to execute an action which would not be in the set \(RB(s)\).
  \item[] - \(DD: \mathcal{S} \rightarrow 2^{\mathcal{A}}\) is a function \(\mathcal{A}^{DD} = DD(s)\) trained on a dataset \(\mathcal{D}\) of (human) trajectories and should approximate the preferences over actions present in the dataset. We expect \(DD\) to help the agent identify tentative actions from within \(\mathcal{A}^{RB}\) that the society that generated \(\mathcal{D} \) prefers.
\end{enumerate}

Thus \(RB\) defines the permitted actions compliant with the mandatory norms while \(DD\) defines actions compliant with the tentative social norms. We expect that for most of the states \(s \in \mathcal{S}\) the two sources of norms \(RB\) and \(DD\) will find some actions in common; that is \(\mathcal{A}^{DD} \cap \mathcal{A}^{RB} \neq \emptyset\). By motivating the agent to select actions from this intersection of safe/legal and socially acceptable actions we hope to find policies which are value aligned. However, it can also be that \(RB\) conflicts with the preferences in \(DD\) and we have \(\mathcal{A}^{DD} \cap \mathcal{A}^{RB} = \emptyset\). This can happen in cases when the society, for whatever reason, does not respect the safety / legal norms. In order to solve conflicts between these two sources we make \(DD\) violable (tentative). However, every time the agent violates the norms in \(DD\) it is penalized. This motivates the agent to find a behaviour which respects both of these normative signals.

Separating the norms into these two components \(RB, DD\) has the following advantages: firstly, it enables us to combine norms represented in different ways. For example, the safety / legal norms can be specified in a logical language as is often done \cite{10.1007/978-3-031-21203-1_5} which makes them verifiable and explainable while the social norms, often considered too difficult to engineer \cite{10.5555/3306127.3331914,10.1145/3278721.3278774}, can be learned through machine learning techniques. Secondly, we keep a clear hierarchy between the norms. By forcing the agent to always choose an action from the permitted set \(\mathcal{A}^{RB}\), HAVA ensures safety / legality of the found policies. In the case of a norm conflict between \(RB\) and \(DD\), the agent is forced to violate the social norms rather than the safety/legal norms.

Our algorithm is controlled by two hyperparameters \(\tau \text{ and } \alpha\). \(\tau\) is used only in the case of continuous action spaces and it is the maximum distance from the norms in \(\mathcal{AV}\) beyond which the agent's reputation goes to zero. \(\alpha\) represents the speed (number of steps) at which the agent's past norm violations are forgiven.

For any state \(s_t \in \mathcal{S}\), the Alignment Value \(\mathcal{AV}\) returns two sets of actions - allowed actions according to the rule-based norms in \( \mathcal{A}_t^{RB} = RB(s_t)\) and allowed actions according to the learned norms in \(\mathcal{A}_t^{DD} = DD(s_t)\). The agent's policy then chooses an action \(a_t = \pi(s_t)\). We proceed to calculate the minimal distances of \(a_t\) to the actions in \(\mathcal{A}^{RB}_t, \mathcal{A}^{DD}_t\) and obtain distances \(d_t^{RB}, d_t^{DD} \geq 0\) respectively. Having obtained the two distances we then calculate how aligned with the rule-based and data-driven norms the action \(a_t\) is by using the Equation \ref{eq:alignment}:
\begin{align}
  \label{eq:alignment}
  al(\tau,d) &= \max \left\{\right .\frac{\tau-d}{\tau}, 0\left.\right\} \\
  & \in [0,1] \nonumber
\end{align}
where \(\tau\) is the maximum distance of the agent's action \(a_t\) from \(\mathcal{A}_t^{RB}\) and \(\mathcal{A}_t^{DD}\). Thus \(\tau\) is a cut-off point beyond which the agent's action is judged as not value aligned at all (\(al(\cdot)=0\)). In other words, \(\tau\) determines how tolerant we are to the agent's mistakes. When \(\tau=0\) then any agent's action that is not in either \(\mathcal{A}^{RB}\) or \(\mathcal{A}^{DD}\) will be judged as misaligned. However, in continuous action spaces it might be useful to tell the agent just \textit{how wrong} it was in order to help the agent distinguish trajectories that were almost aligned from those that were not.

Using the Equation \ref{eq:alignment} we compute how aligned the agent's action was to the norms in \(RB\): \(al_t^{RB}=al(\tau,d_t^{RB}), \text{ and }DD \text{: } al_t^{DD}=al(\tau,d_t^{DD})\) and we find the smaller of the two quantities: \(\delta_t = \min \left\{al_t^{RB}, al_t^{DD}\right\}\). Thus \(\delta_t \in [0,1]\) represents the worst of the two alignments at time \(t\).

This is then put into a context of the agent's past behaviour (its reputation for following the norms) summarized by \(w_t \in [0,1]\):
\begin{align}
  w_{\text{inc}}(w_t) & = \alpha \left[e^{w_t}-1\right]+0.001  \label{eq:weight_increment}\\
  w_{t+1} &= \min \left\{w_t + w_{\text{inc}}(w_t), \delta_t\right\}\label{eq:weight_of_V}
\end{align}
The Equation \ref{eq:weight_of_V} is in charge of computing the agent's reputation for the next time-step. When the agent violates the norms the reputation falls to \(\delta_t \in [0,1)\). On the other hand, when the agent follows all the norms its reputation \(w_{t+1}\) either remains \(1\) or is increased towards \(\delta_t = 1\).

The Equation \ref{eq:weight_increment} is the step by which \(w_t\) increases in the next time-step if the agent follows the norms. We have chosen this non-linear equation as a metaphor for forgiveness. The function rises slower the lower the agent's reputation \(w_t\) is. As a result, an agent which severely violated the norms (\(d^{RB}_t \geq \tau\) or \(d^{DD}_t \geq \tau\)) will take longer to recover the reputation \(w_t\) back to 1 than an agent whose violation was lower (\(d^{RB}_t < \tau\) or \(d^{DD}_t < \tau\)). How quickly the equation rises is decided by \(\alpha \in [0,\infty)\). To see how \(\alpha\) controls the reputation growth consider the following example: let \(w_t=0, \alpha=10\) and assume that the agent's next four actions are all value aligned (\(\delta_{t+1} =\delta_{t+2} =\delta_{t+3} =\delta_{t+4} = 1\)). Then \(w_t\) will be updated as follows: \(w_{t+1}=0.001, w_{t+2}\approx 0.012,w_{t+3}\approx 0.141,w_{t+4}=1\). In other words, it will take the agent just 4 aligned actions to recover its reputation \(w\) back to 1.

\begin{figure}[t]
  \centering
  \includegraphics[width=7cm]{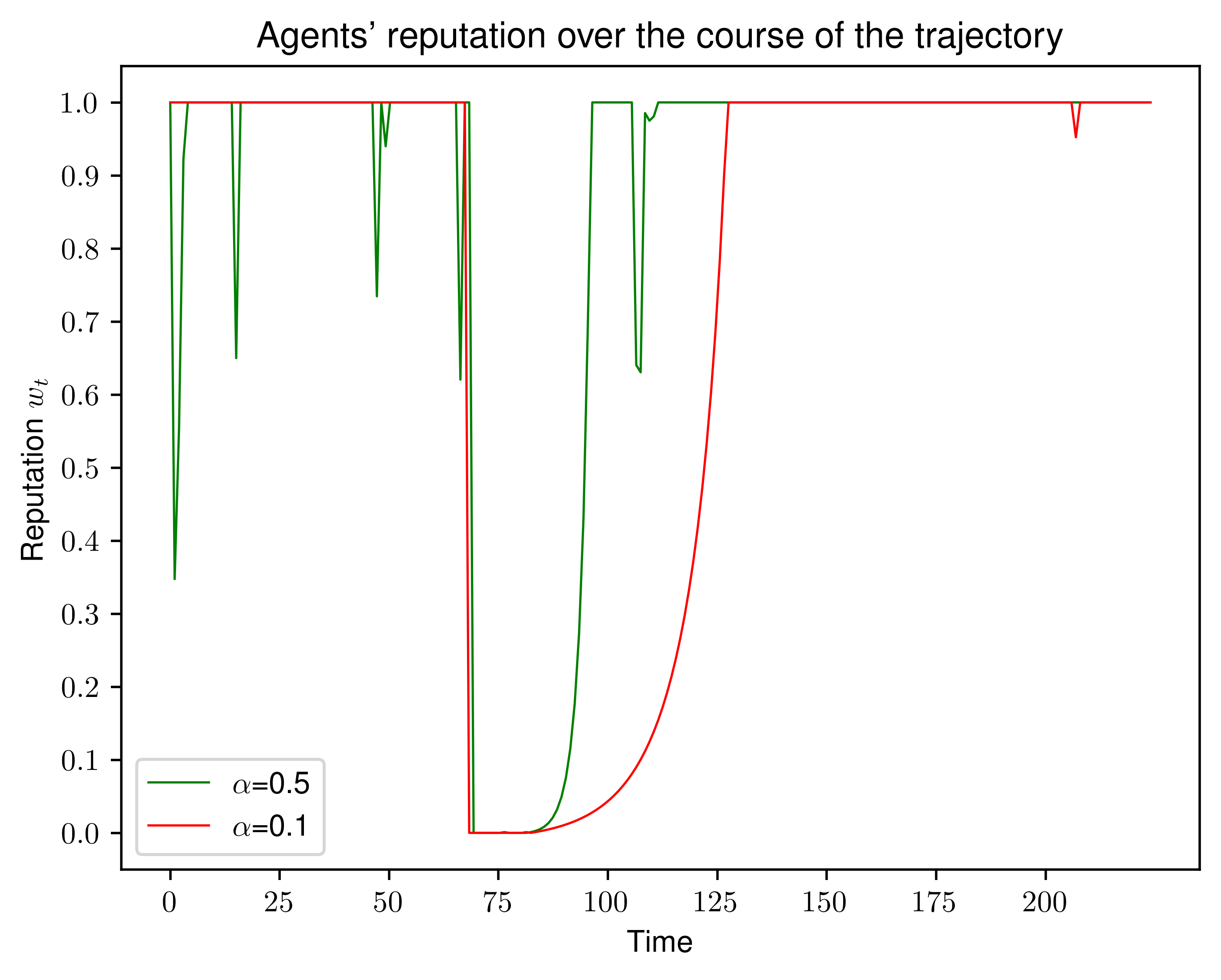}
  \caption{Development of two HAVA agents' reputation \(w_t\) in the junction scenario. We can see that \(\alpha=0.5\) (15 steps) tends to violate the norms more often (more dips) than \(\alpha=0.1\) (45 steps) because it is forgiven faster. This can be seen around the time \(t=75\) when both agents violate the \(DD\) norms by more than \(\tau=1\) and their reputation decreases to 0. The agent with \(\alpha=0.5\) recovers its reputation much quicker than the agent with \(\alpha=0.1\).}
  \label{fig:inc_and_forgiveness}
\end{figure}

In Figure \ref{fig:inc_and_forgiveness}, we see that as \(\alpha\) decreases to 0, the number of steps it takes for the agent to recover its reputation increases - when \(\alpha=0.5\) it takes 15 steps, \(\alpha=0.1\) takes 45 steps etc. Thus \(\alpha\) can have a profound impact on what policies are considered value aligned and therefore optimal.

\subsection{Embedding Values into the MDP}
\label{sub:embedding_values_into_the_mdp}

Given an MDP \(\mathcal{M}=\left\langle \mathcal{S}, \mathcal{A}, R_{\text{task}}, T\right\rangle\) and an Alignment Value \(\mathcal{AV}=\left\langle RB, DD\right\rangle\) we consider the agent's policy \(\pi\) to be value aligned with respect to \(\mathcal{AV}\) if and only if:
\begin{enumerate}
  \item[] - its produced trajectories \(\sigma\) are statistically indistinguishable (p-value > 0.05) from the dataset \(\mathcal{D}\) for any states \(s \in \mathcal{S}\) where \(RB(s) \cap DD(s) \neq \emptyset\)
  \item[] - it complies with all norms in \(RB(s)\) for any \(s \in \mathcal{S}\) where \(RB(s) \cap DD(s) = \emptyset\)
\end{enumerate}
In order to achieve this, we modify the original MDP \(\mathcal{M}\) in two ways:
\begin{enumerate}
  \item add the current agent's reputation \(w_t\) into the state space \(\mathcal{S}\), and
  \item weigh the task reward \(R_{\text{task}}\) with \(w_t\).
\end{enumerate}
At a time-step \(t\), the environment presents a new state \(s_t\). The agent then chooses an action \(a_t\) based on the state \(s_t\) and the current reputation: \(a_{t} = \pi(s_t \cup w_t)\). This action is checked against the Alignment Value \(\mathcal{AV}\) and a weight \(w_{t+1}\) is computed based on Equation \ref{eq:weight_of_V}. If \(a_t\) complies with the rule-based norms in \(\mathcal{A}^{RB}\), that is \(a_t \in \mathcal{A}^{RB}\), \(a_t\) is executed otherwise an action from \(\mathcal{A}^{RB}\) is selected instead.

Upon executing the action, the environment returns a reward \(r_{t} = R_{\text{task}}(s_t,a_t)\) corresponding to the transition \(s_t, a_t\). We weigh this reward by the agent's reputation \(w_{t+1}\) obtaining \(R_{\mathcal{AV}}(s_t,a_t,w_{t+1})\) as shown in Equation \ref{eq:reward_weighted}:
\begin{align}
  \label{eq:reward_weighted}
  \text{Let }r_{t} &= R_{\text{task}}(s_t,a_t) \text{ then} \nonumber\\
  R_{\mathcal{AV}}(s_{t}, a_{t},w_{t+1}) &= \left\{
    \begin{array}{rcl}
      w_{t+1} r_{t} & \text{ if } r_{t} \geq 0 \\
      r_{t} \left[1+(1-w_{t+1})\right] & \text{ if } r_{t} < 0
    \end{array}
  \right.
\end{align}
Thus the agent can receive the original task reward \(r_{t}\) only if it has been respecting the Alignment Value \(\mathcal{AV}\). Thus relying on the reputation \(w_t\) enables HAVA to side-step the difficult task of hand engineering a value aligned reward function.

Thus, given an MDP \(\mathcal{M}\) and an Alignment Value \(\mathcal{AV}\), we obtain a new MDP \(\mathcal{M}'=\left\langle \mathcal{S}', \mathcal{A}, R_{\mathcal{AV}}, s_0, T\right\rangle\) where \(\mathcal{S}'\) is an augmented state space \(\mathcal{S}' = \mathcal{S} \cup [0,1]\) and \(R_{\mathcal{AV}}\) is the weighted task reward function \(R_{\text{task}}\) as seen in Equation \ref{eq:reward_weighted}. The policy is newly defined as \(\pi: \mathcal{S}' \rightarrow \mathcal{A}\) and it maximizes the following objective \(\mathbb{E}_{\sigma\sim \pi} \left[\right. \textstyle\sum_{t=0}^{\infty}\gamma^t R_{\mathcal{AV}}(s_t, a_t, w_{t+1})\left.\right]\). The diagram of our architecture is given in Figure \ref{fig:architecture}.


\section{Experiments}


\subsection{Toy Example: Grid World}
\label{sub:toy_example:_grid_world_mdp}

\begin{figure}[t]
  \centering
  \includegraphics[width=4cm]{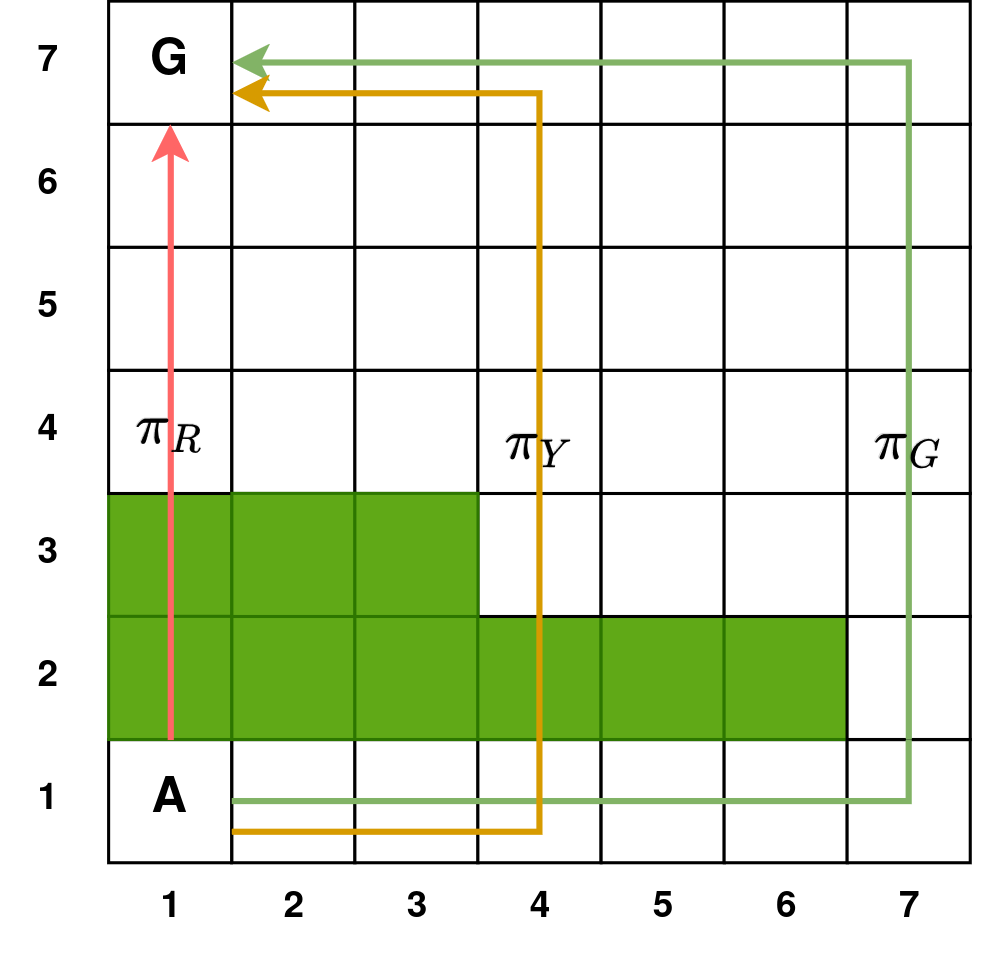}
  \caption{Toy Example: Grid world with three policies \(\pi_R, \pi_Y, \pi_G\). The social norms in \(DD\) prohibit visiting the lawn tiles (green). The expected discounted return of the offending policies \(\pi_R, \pi_Y\) depends on \(\alpha\) while the return of \(\pi_G\) remains unchanged.}
  \label{fig:toymdp}
\end{figure}

In this toy example, shown in Figure \ref{fig:toymdp}, we want to show how HAVA motivates the agent to converge to a value aligned policy, by changing the values of \(\alpha\). We are given a discrete MDP \(\mathcal{M} = \left\langle \mathcal{S}, \mathcal{A}, T, R_{\text{task}}\right\rangle\) where there are 49 states including 9 lawn tiles (green), 4 actions (up, down, left, right). The agent maximizes the following reward function
\begin{align*}
  R_{\text{task}}(s_t,a_t) &= \left\{
    \begin{array}{rcl}
      -1 & \text{ otherwise}\\
      100 & \text{ when on goal } G
    \end{array}
  \right.
\end{align*}
We employ HAVA with its Alignment Value \(\mathcal{AV}=\left\langle RB, DD\right\rangle\) where we assume the mandatory norms in \(RB\) forbid the agent from leaving the bounds of the MDP while the tentative social norms in \(DD\) dictate to avoid the lawn tiles. Since this is a discrete action space we do not need to use \(\tau\) here and assume that whenever the agent violates the norms its reputation \(w=0\).

For this example we generated three policies that reach the goal (G) in different ways. Two of them, \(\pi_R, \pi_Y\), violate the social norms by passing through the lawn while the third policy \(\pi_G\) avoids it altogether. We would like to show how violating the social norms, becomes less and less attractive for the agent as \(\alpha\) decreases.
In HAVA, every policy generates trajectories such as \(\sigma\):
\begin{align*}
  \sigma &= \left(s_t, w_t, a_t, s_{t+1}, w_{t+1}, r_{t}, \dots \right)
\end{align*}
Here we give an example of computing the discounted return for the policy \(\pi_R\) with \(\alpha=10\) (4 steps) and \(\alpha=5\) (5 steps):
\begin{align*}
  \sigma_{\pi_R} &= \left(s_{11}, 1, a_{\text{up}}, s_{1,2}, 0, -2, \dots \right) \text{ // \(\pi_R\) trajectory }\\
  J_{\alpha=10}(\pi_R) &= \mathbb{E}_{\sigma_{\pi_R}\sim \pi_R} \left[\right. \textstyle\sum_{t=0}^{\infty}\gamma^t R_{\mathcal{AV}}(s_t, a_t, w_{t+1})\left.\right] \\
  &= -2 -2 \gamma^1 -2 \gamma^2 -1.99 \gamma^3 - 1.87 \gamma^4 + 100 \gamma^5 \approx 86 \\
  J_{\alpha=5}(\pi_R) &= \mathbb{E}_{\sigma_{\pi_R}\sim \pi_R} \left[\right. \textstyle\sum_{t=0}^{\infty}\gamma^t R_{\mathcal{AV}}(s_t, a_t, w_{t+1})\left.\right] \\
  &= -2 -2 \gamma^1 -2 \gamma^2 -1.99 \gamma^3 - 1.96 \gamma^4 + 26 \gamma^5 \approx 15
\end{align*}
In this way we calculate the expected returns for the remaining policies and \(\alpha\). The resulting discounted returns are presented in Table \ref{tab:toy_results}. We notice that while \(\alpha=10\) allowed the agent to recover its reputation and receive \(100 \gamma^5\) reward at the end, with \(\alpha=5\) this was not the case and the agent only receives \(w_5 100 \gamma^5 = 0.26 * 100 * 0.99^5 = 24.72\).

\begin{table}[H]
  \centering
  \begin{tabular}{c | c | c | c}
    \hline
    \(\alpha\) & \(J(\pi_R)\) & \(J(\pi_Y)\) & \(J(\pi_G)\) \\
    \hline
    10 (4 steps) & \textbf{86} & 76 & 67\\
    5 (5 steps) & 15 & \textbf{75} & 67\\
    4 (6 steps) & 5 & \textbf{75} & 67\\
    2 (7 steps) & -5 & \textbf{73} & 67\\
    1.6 (8 steps) & -7 & \textbf{73} & 67\\
    1.2 (9 steps) & -8 & 27 & \textbf{67} \\
    1 (10 steps) & -8 & 6 & \textbf{67} \\
    \hline
  \end{tabular}
  \caption{As \(\alpha\) decreases, the preference shifts from the most offending \(\pi_R\) to less offending \(\pi_Y\) to norm satisfying \(\pi_G\). For \(\alpha\leq 1.2\), \(\pi_G\) will always be optimal in this MDP.}
  \label{tab:toy_results}
\end{table}

We notice that whichever policy is considered optimal changes as we increase the number of steps necessary to recover the lost reputation \(w\). A very low value of \(\alpha\) will punish even the least serious offences while higher values of \(\alpha\) might allow for some forms of social norms violations but not others (such as \(\pi_Y\)).

It is interesting to note that decreasing the value of \(\alpha\) does not effect the discounted return of the value aligned policies such as \(\pi_G\) as can be seen in Table \ref{tab:toy_results}. Therefore, once we have found an \(\alpha\) for which the value aligned policy becomes optimal, we can stop the search because any lower values of \(\alpha\) will not change the optimality of \(\pi_G\). This can serve as a useful method for finding the correct \(\alpha\).

\subsection{Junction Scenario}
\label{sub:junction_scenario}

\begin{figure}[H]
  \centering
  \includegraphics[width=6cm]{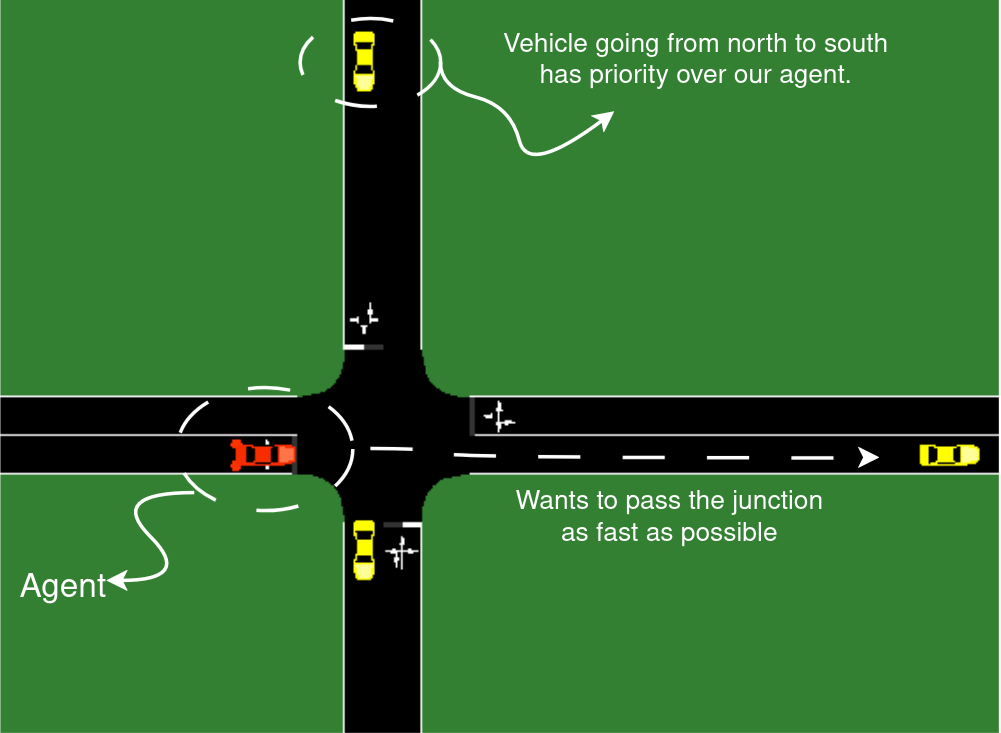}
  \caption{The agent is spawned and must learn how to cross the junction in the fastest way while respecting the necessary safety, legal and social norms.}
  \label{fig:task}
\end{figure}

We test HAVA in a traffic simulator SUMO \cite{SUMO2018}. The agent is spawned on the west road and needs to learn how to drive through a busy junction to exit on the east. A vehicle going from north to south has a higher priority than our agent. The scenario is visualized in Figure \ref{fig:task}. In the simulator it takes the agent on average about 22 seconds during which it chooses 220 actions (speeds).

The task reward the agent is maximizing in this environment is defined as:
\begin{align*}
  R_{\text{task}}(s_t, a_{t-2},a_{t-1},a_t) &= \left\{
    \begin{array}{rcl}
      10 + \frac{a_{t-2} + a_{t-1} + a_t}{3} & \text{ every 2 meters}\\
      -1 & \text{ otherwise}
    \end{array}
  \right.
\end{align*}
This reward function rewards the agent every two meters based on the last three actions (speeds). This encourages the agent to pass the junction as quickly as possible and also means the agent is motivated to violate any norms that slow it down.

This is a continuous state space environment and therefore, in our experiments, we use a deep Q-learning \cite{Mnih2015} with several improvements, namely: double \cite{vanhasselt2015deepreinforcementlearningdouble}, duelling \cite{10.5555/3045390.3045601} and noisy networks \cite{fortunato2019noisynetworksexploration}. We discretized the otherwise continuous action space into 11 actions through which the agent controls the speed of the vehicle, namely: we let the agent choose between 5 accelerating actions which increase the speed by some fixed amount, 5 decelerating actions which decrease it and 1 action that maintains the current speed.

\subsubsection{HAVA}
\label{subsub:hava}

HAVA uses an Alignment Value \(\mathcal{AV}=\left\langle RB, DD\right\rangle\) which comprises of two sources of norms \(RB\) and \(DD\). In this driving scenario, norms in \(RB\) represent the speed limits (50km/h) and the rules of the junction (e.g. which vehicle has priority over others). On the other hand, the social norms in \(DD\) represent the acceleration and deceleration of vehicles. In our experiments we show why both \(RB\) and \(DD\) are important and how through varying \(\alpha\) we can generate policies which are value aligned.

To implement the rule-based norms in \(RB\) we use the SUMO's Krauss model \cite{Krauss1998, SUMO2018}. Under the Krauss model, the vehicles in the simulation maintain as high speed as possible without causing an accident. This means that if our agent attempts to travel at a dangerous speed the simulator will execute the closest safe action instead. On top of the Krauss model, we manually added a speed limit norm forbidding the agent from travelling faster than 50km/h.

\begin{figure}[t]
  \centering
  \includegraphics[width=7cm]{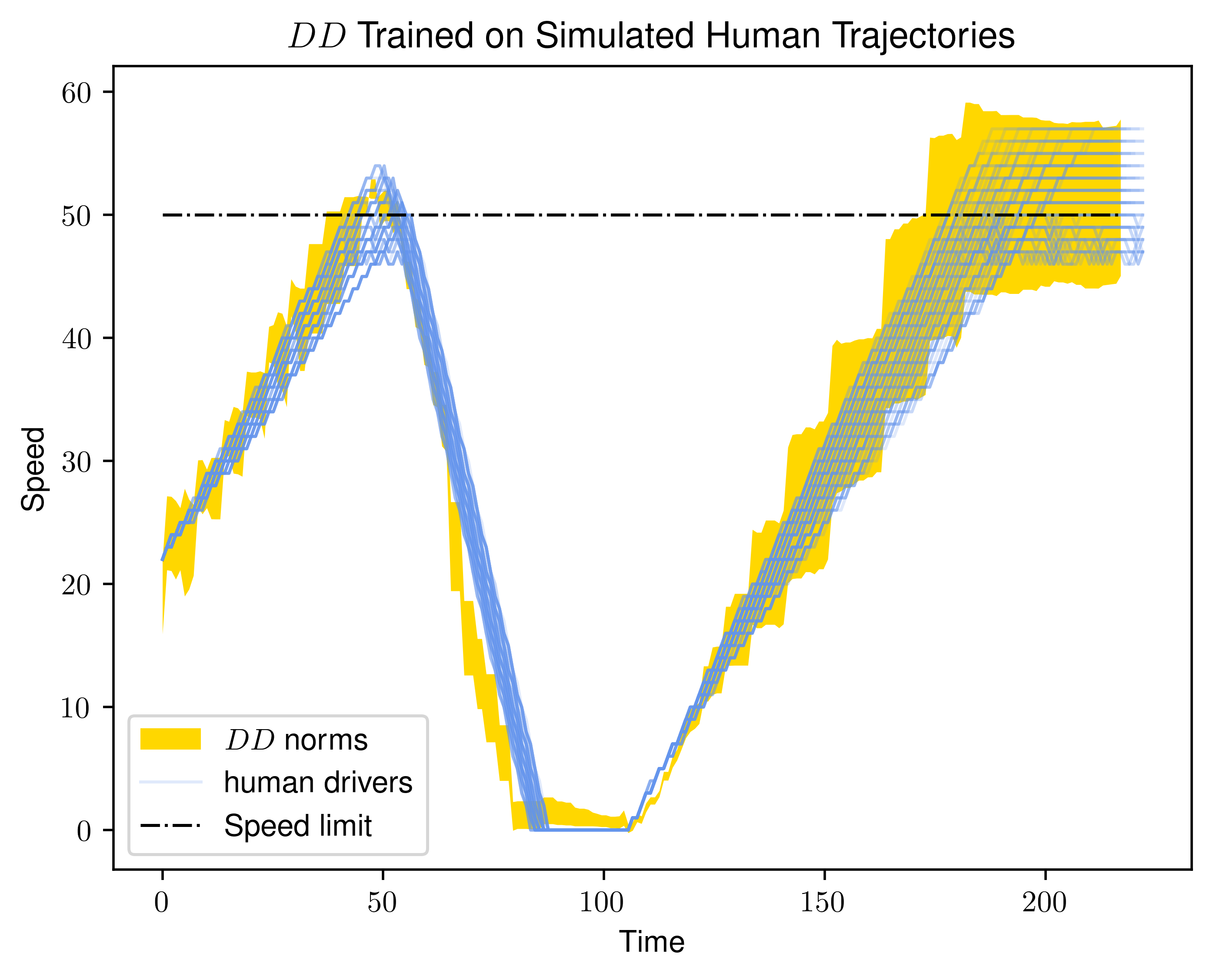}
  \caption{Simulated human trajectories used to train \(DD\) to predict the social norm of speed. The human drivers accelerate (time \(t=50\)), occasionally violating the speed limit, then stop at the junction to let the north-south vehicle pass (time \(t=100\)) and finally accelerate to leave the junction (time \(t=200\)).}
  \label{fig:human_passes}
\end{figure}

To implement the social norms captured in \(DD\) we collected a dataset of simulated human behaviours. These human behaviours follow the Krauss model at different maximal speeds and accelerations. This way we managed to obtain a diverse dataset of behaviours that mimic the social norms (accelerations / decelerations) of a population of drivers. We used supervised learning to train \(DD\) to solve a regression problem of predicting a minimal and maximal speed. The simulated human trajectories used for training as well as the resulting \(DD\) are visualized in Figure \ref{fig:human_passes}.

Thus, whenever the agent selects an action \(a^\pi_t\), the simulator executes an action \(a^{RB}_t\). We can measure the distance between the two as \(d_t^{RB} = |a^\pi_t - a^{RB}_t|\). If \(a^\pi_t\) respects the Krauss model the distance will be 0. The \(DD\) model returns a range of speeds and the distance is calculated as \(d_t^{DD} = \min \left\{|a^\pi_t-a^{DD}_t| \text{ s.t. } a^{DD}_t \in \mathcal{A}^{DD}\right\}\).

HAVA is controlled by two hyperparameters \(\tau\) and \(\alpha\). \(\tau\) describes how tolerant we are to the agent's violations of the norms. \citet{10.1145/3278721.3278728} propose that artificial agents be evaluated to the same standards as humans when it comes to morality. We employ this view and use it to set \(\tau=1\). This means that whenever the agent diverges from the social norms by more than 1km/h its reputation \(w_{t+1}=0\). We believe that potentially violating the social norms up to 1km/h in this scenario would be acceptable for human drivers as well.

The hyperparameter \(\alpha\) controls how quickly the reputation \(w_t\) can recover. In other words, how many consecutive norm-respecting actions it takes for \(w_t\) to recover from 0 back to 1. \(\alpha=0.5\) corresponds to 15 actions while \(\alpha=0.1\) to 45 actions as visualized in Figure \ref{fig:inc_and_forgiveness}. An agent whose reputation \(w_t < 1\) receives lower task rewards than a value aligned agent. We expect this to have a significant impact on the found policies.

The goal of this experiment is to show that HAVA is capable of mixing the rule-based and data-driven norms and achieving a value aligned performance even in the case of continuous Markov decision processes.

\subsubsection{Ablation: Rule-Based Norms and RL Only}
\label{subsub:rule-based_norms_and_rl_only}

We carry out the same experiment as with HAVA but this time only specify the rule-based norms. Thus the agent is given an Alignment Value \(\mathcal{AV}^{RB} = \left\langle RB\right\rangle\). We run this experiment with several values of \(\alpha = \left\{10,0.5,0.1\right\}\) and \(\tau=\left\{1\right\}\). The agent is not allowed to violate the rule-based norms.

The goal of the experiment is to show the importance of the social norms learned from data. We assume, as is done elsewhere in the literature \cite{10.1145/3278721.3278774}, that some norms, such as the social norms, cannot be efficiently encoded using a rule-based approach due to the number of such norms. Thus \(RB\) only contains the rule-based safety and legal norms.

\subsubsection{Ablation: Data-Driven Norms and RL Only}
\label{subsub:data-driven_norms_and_rl_only}

Our second ablation considers a purely data-driven approach with Alignment Value \(\mathcal{AV}^{DD} = \left\langle DD\right\rangle\). Similarly, we run this experiment with several values of \(\alpha = \left\{10,0.5,0.1\right\}\) and \(\tau=\left\{1\right\}\).

Importantly, the agent is allowed to violate the learned norms (SUMO does not enforce any safety rules and crash detection is off). The goal is to demonstrate how the lack of understanding of norm severity can cause safety / legal norm violations. The data-driven approaches are typically unable to tell apart the tentative norms that can be occasionally violated (e.g. social norms) from the mandatory norms that cannot (e.g. safety norms). In this experiment we would like to show that, given a choice, a data-driven method might choose to violate safety critical norms to maximize its task rewards. On the other hand, our approach HAVA will always choose to violate the social norms rather than the safety / legal norms.

Not understanding the norm severity is just a single problem that the data-driven approaches face. Other problems typically include the need for a perfectly aligned dataset, not being able to learn temporally extended norms \cite{osti_10301363} and a lack of interpratibility.
\begin{table}[t]
  \centering
  \begin{tabular}{| c | c | c | c |}
    \hline
     & RB & HAVA & DD \\
    \hline
     RB & 1 & & \\
     HAVA & \(2.4*10^{-180}\) & 1 &  \\
     DD & \(9.5*10^{-36}\) & \(1.7*10^{-238}\) & 1 \\
     Human & \(3.3*10^{-239}\) & \(\mathbf{0.42}\) & \(1.2*10^{-321}\) \\
    \hline
  \end{tabular}
  \caption{2-sample KS test proves that the policy found by HAVA is statistically indistinguishable from the human trajectories (p-value = 0.42).}
  \label{tab:ks_2samp_mixhuman_tau1}
\end{table}

\begin{figure*}
  \begin{minipage}{.45\linewidth}
    \centering\includegraphics[width=0.7\linewidth]{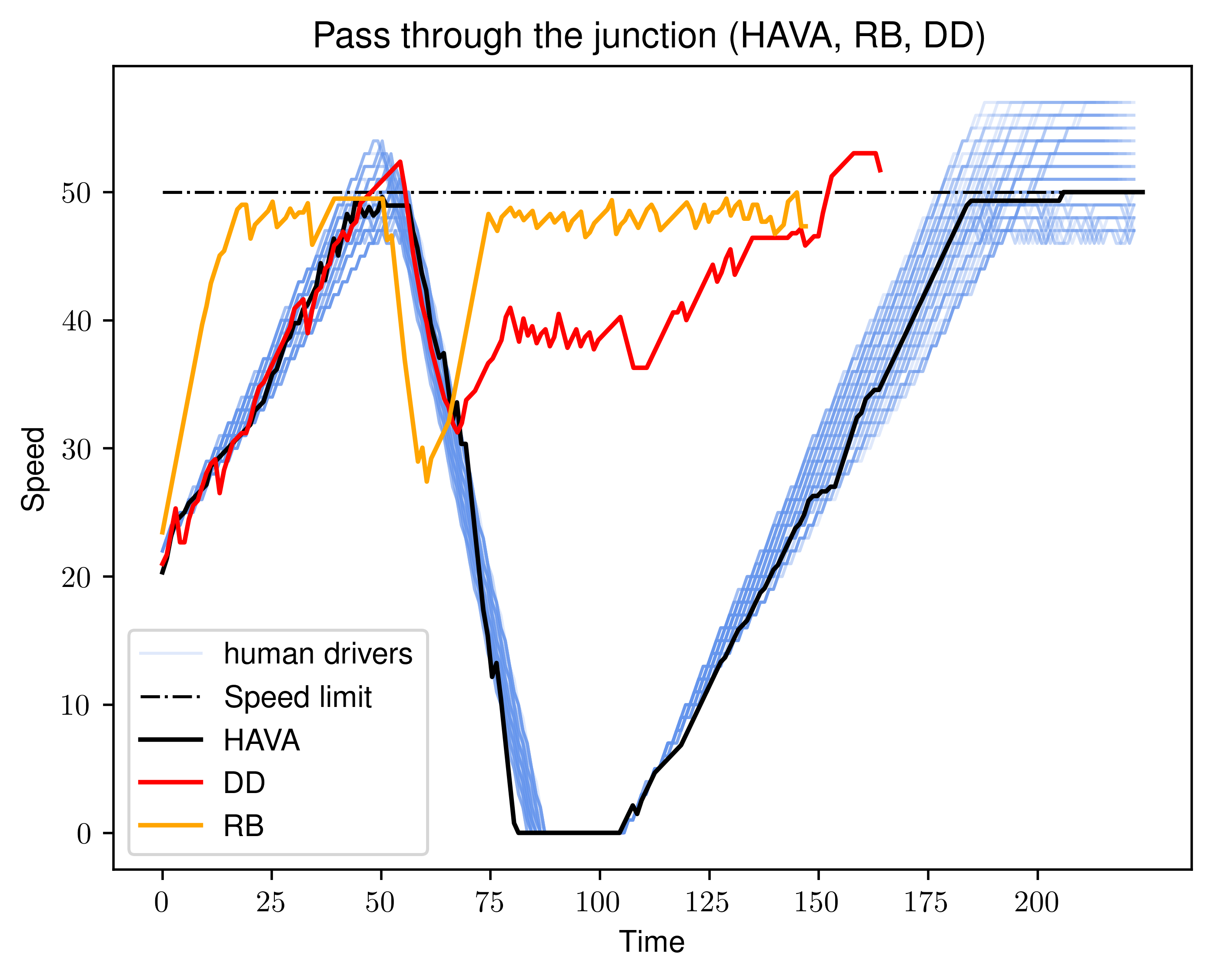}
    \captionof{figure}{We see the three strategies that the pure data-driven (DD), rule-based (RB) and HAVA approaches converged to. All three of these had \(\alpha=0.1\). We can see that only HAVA was able to find a policy that resembled the human behaviours (blue).}
    \label{fig:pass_together_01}
  \end{minipage}
  \hfill
  \begin{minipage}{.45\linewidth}
    \centering\includegraphics[width=0.7\linewidth]{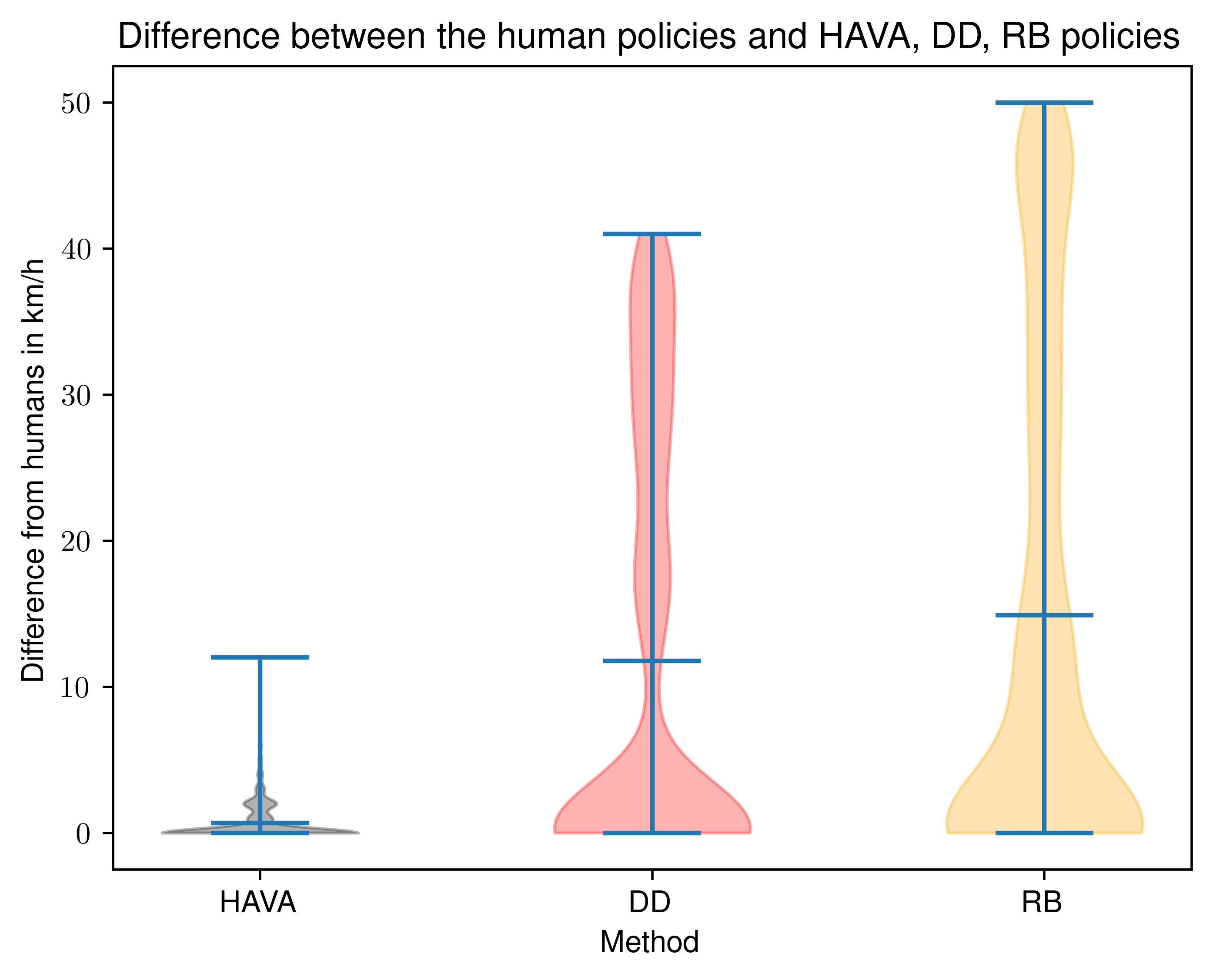}
    \captionof{figure}{Here we see the differences between the human trajectories and the data-driven, rule-based and HAVA trajectories. HAVA clearly violates the human trajectories the least with median violation of 0 and mean violation of 0.8 km/h.}
    \label{fig:differences_together_01}
  \end{minipage}

  \vspace*{\floatsep}

  \begin{minipage}{.3\linewidth}
    \centering\includegraphics[width=1.1\linewidth]{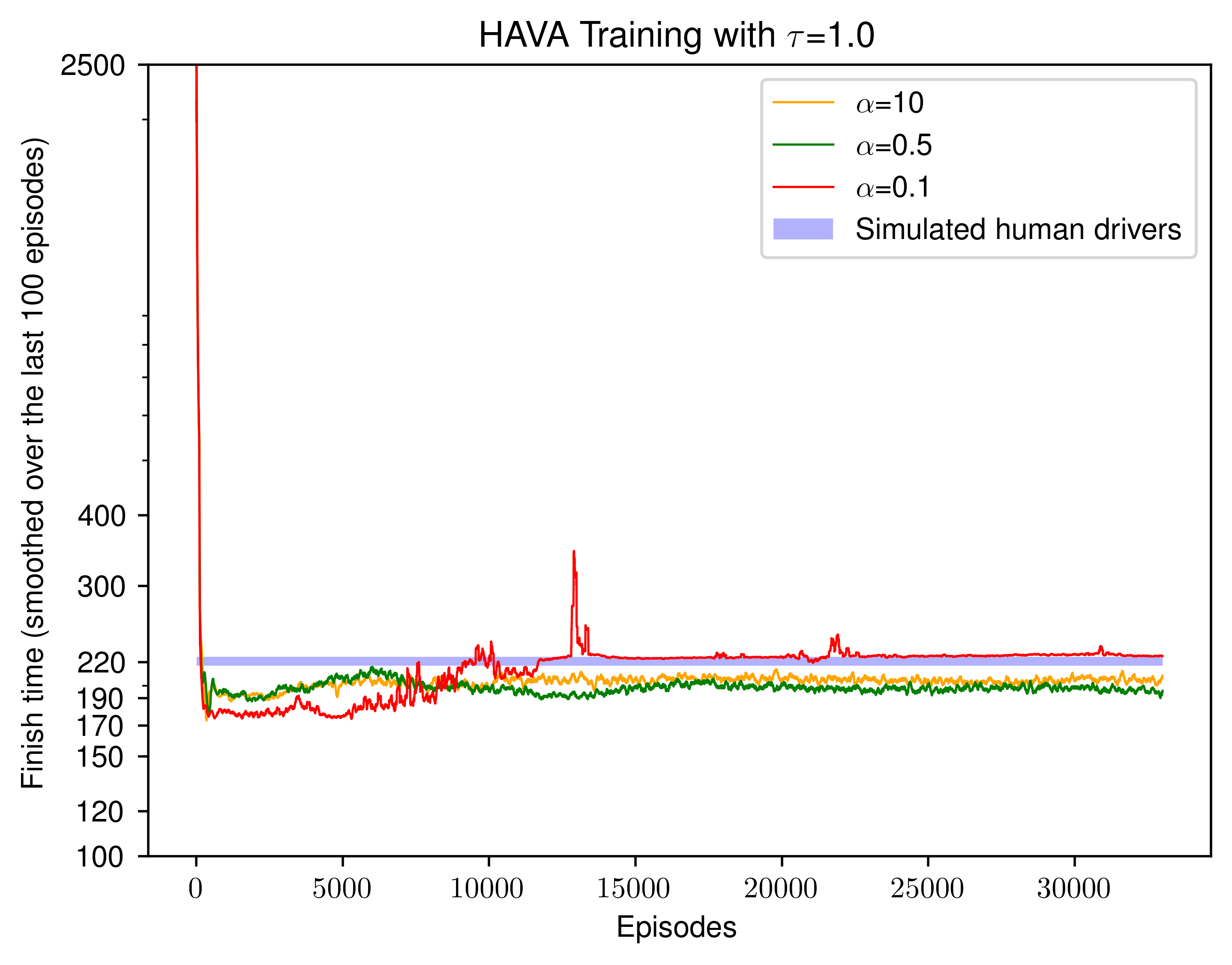}
    \captionof{figure}{HAVA produces two clusters of policies. While \(\alpha\) of 10 (4 steps) and 0.5 (15 steps) does not seem to be able to converge to a human-like behaviour, setting \(\mathbf{\alpha=0.1}\) (45 steps) produces a value aligned policy.}
    \label{fig:mixtraining}
  \end{minipage}
  \hfill
  \begin{minipage}{.3\linewidth}
    \centering\includegraphics[width=1.1\linewidth]{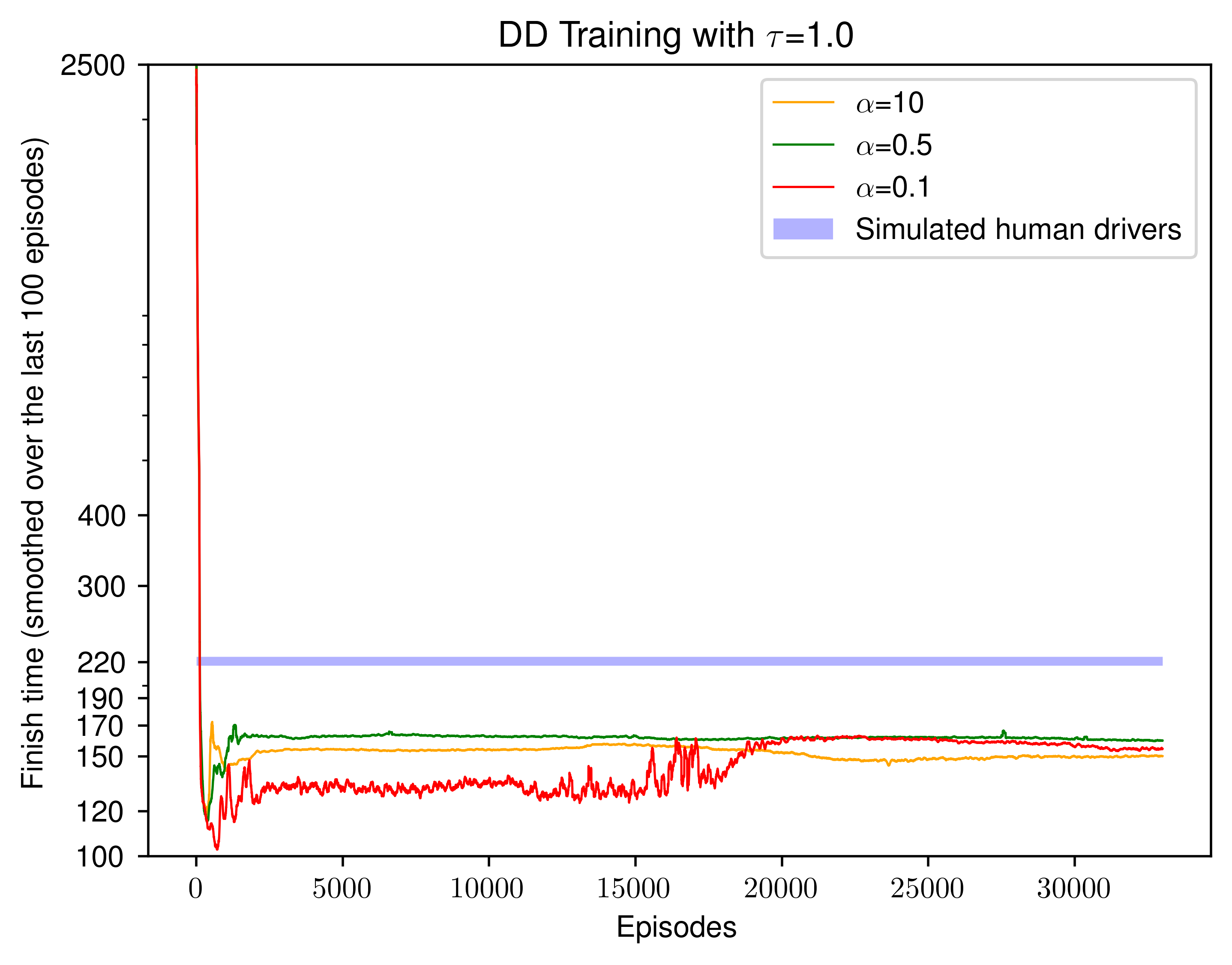}
    \captionof{figure}{The pure data-driven method produces policies which violate the safety norms and pass the junction without stopping. This highlights the data-driven methods' problem of not distinguishing the norms's severity.}
    \label{fig:ddtraining}
  \end{minipage}
  \hfill
  \begin{minipage}{.3\linewidth}
    \centering\includegraphics[width=1.1\linewidth]{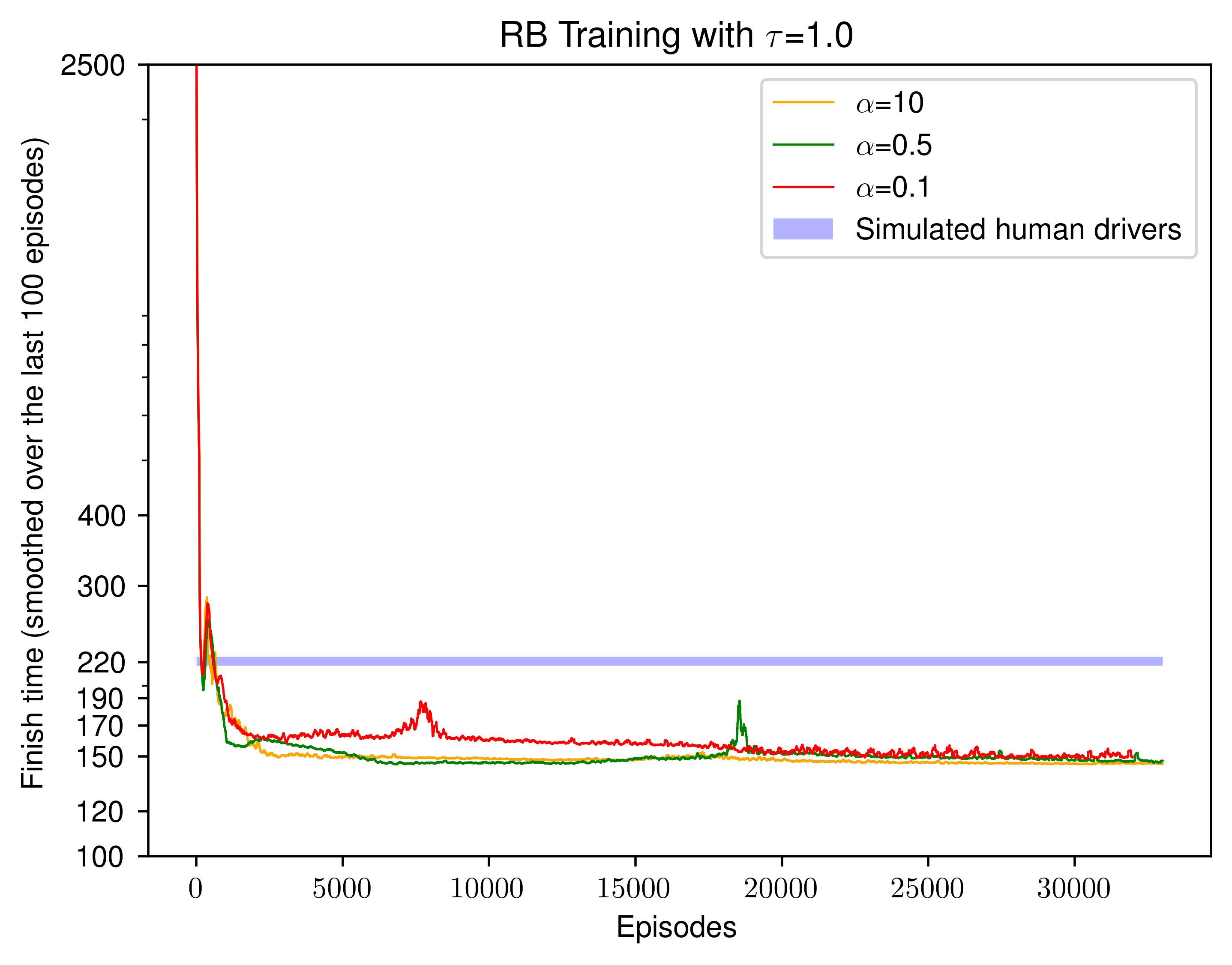}
    \captionof{figure}{The rule-based approach finds policies which respect the safety/legal norms but the lack of social norms enables the agent to execute some undesirable actions such as aggressive acceleration.}
    \label{fig:rbtraining}
  \end{minipage}
\end{figure*}%

\subsubsection{Results: HAVA}
\label{subsub:results_hava}

In this experiment we wanted to demonstrate that our proposed method, HAVA, produces value aligned behaviours. Similarly to the toy example we try out several values of \(\alpha\) before arriving at a value aligned policy. Relatively high values of \(\alpha=10,0.5\) do not seem to affect the agent sufficiently to force it to respect the necessary norms as seen in Figure \ref{fig:mixtraining}.

With lower value of \(\alpha=0.1\) we observe the agent's policy resembles that of a human driver (e.g. waits at the junction for a similar amount of time). We have trained the agent for about 35K episodes (Figure \ref{fig:mixtraining}). Besides the training curves of \(\alpha=10,0.5,0.1\) we also plot the range of human policies (in blue). The humans finish passing the junction between times 217-225. We would expect a value aligned policy to fall within this range. As we can see the policies for \(\alpha=10,0.5\) seem to finish on average around time 200 which is too fast. However, when setting \(\alpha=0.1\) (45 steps) we relatively quickly obtain a value aligned behaviour where the average finish time is 225.

This policy is also statistically indistinguishable from the human behaviours according to the 2-sample KS test as seen in Table \ref{tab:ks_2samp_mixhuman_tau1}. Finally, in Figure \ref{fig:differences_together_01} we can see the difference (in km/h) between the human and HAVA trajectories. We see that the found policy has a median violation of the social norms of 0 and mean of 0.8km/h. For the KS test and the difference plots we used 10 policies from the last 500 training episodes.
\subsubsection{Results: Rule-Based and RL}
\label{subsub:rulesonly}

In this experiment we wanted to demonstrate the importance of the tentative social norms. When we use RL to optimize for \(\mathcal{AV}^{RB}\), the agent learns to respect all the safety and legal norms but nevertheless the resulting policy is not value aligned as seen in Figure \ref{fig:pass_together_01}. The reason for this is that the human behaviour is shaped by more than laws and safety rules. Human societies also develop many social norms, unwritten rules, which develop and change over time. These are also the hardest to define manually and currently the literature seems to assume these norms have to be learned.

The trained policies shown in Figure \ref{fig:rbtraining} converge to a blazing finish time of 150 timesteps. This is much faster passage than that of the human drivers (217-225 time-steps). The reason for this is that while the norms in \(RB\) are concerned with safety / legality they do not attempt to define what a socially acceptable acceleration is. In Figure \ref{fig:pass_together_01}, we can see what such policies look like. The lack of social norms, turned the agent into an aggressive driver who accelerates very abruptly (around time 25 in Figure \ref{fig:pass_together_01}) in order to avoid stopping and waiting at the junction. We have carried out the 2-sample KS test (Table \ref{tab:ks_2samp_mixhuman_tau1}) and with p-values of \(3.3 * 10^{-239}\) we conclude that these policies do not resemble any of the HAVA or human policies. Although this strategy is safe according to the rules in SUMO, it is nevertheless unacceptable. Therefore, if we assume that not all norms can be manually encoded within the agent, some hybrid solution such as HAVA will be necessary if we want to compute value aligned policies.

\subsubsection{Results: Data-Driven and RL}
\label{subsub:data_driven_and_rl}

As shown in Figure \ref{fig:pass_together_01}, the acceleration of the pure data-driven agent resemble the human behaviour much more (around time 25 in Figure \ref{fig:pass_together_01}) than in the case of the pure rule-based agent. In fact, the data-driven policy even started speeding similarly to the humans in our dataset (around time 50 in Figure \ref{fig:pass_together_01}). This is a common characteristic of the data-driven approaches that they are sensitive to any misaligned data in the dataset and could potentially learn some incorrect norms. This can be solved if we have a perfectly aligned dataset, however as noted by \cite{10.5555/3504035.3504241} the data may not be always available.

There is, however a more serious problem with the data-driven approaches; the inability to distinguish which norms are mandatory (i.e. not understanding the severity of norms). From the training graph (Figure \ref{fig:ddtraining}) we see the policies converged to finish times of around 160 time-steps. This is a bit slower than in the case of the pure rule-based approach. The reason for this are the social norms present in \(DD\). Thanks to \(DD\) the agent accelerates at a similar rate as any human driver. However, as the agent slows down (around time \(t=75\) in Figure \ref{fig:pass_together_01}) it suddenly decides to pass the junction. At this moment the agent violates the safety norms and behaves in a way that could endanger other vehicles at the junction. This can be seen in the demo.

As we can see in Figure \ref{fig:pass_together_01}, the data-driven agent's policies resemble neither the HAVA agent nor the rule-based agent. We have carried out the 2-sample KS test (Table \ref{tab:ks_2samp_mixhuman_tau1}) to prove it. The reason for this is that the data-driven agent is capable of violating the safety and legal norms and therefore can produce a broader set of policies than either HAVA or a pure rule-based approach. The policy when \(\alpha=0.1\) is an example of this as the agent passes the junction by violating the safety norms.

To summarize, having an Alignment Value \(\mathcal{AV}\) which contains two components of norms, one violable and one non-violable is useful. When the Alignment Value \(\mathcal{AV}\) only contains the rule-based norms, the resulting policies will be unsocial. On the other hand, if the Alignment Value \(\mathcal{AV}\) only contains the learned norms in \(DD\), the resulting policies may break norms which are safety critical and therefore become dangerous.

HAVA never violates the rule-based norms in \(RB\) and is motivated by the agent's reputation \(w_t\) to find such a path through the environment that respects the maximum amount of the social norms in \(DD\) as well.


\section{Discussion}
\label{sec:discussion}

\textbf{Related work:} Approaches that learn norms from data often mix together the tentative (social) and mandatory (safe/legal) norms \cite{10.5555/3504035.3504241,ijcai2019p891,DBLP:journals/corr/abs-2104-09469}. This then makes it difficult to guarantee the agent is going to respect the latter. HAVA addresses this by relying on two separate components 1) an interpretable rule-based component (\(RB\)) which enforces the mandatory norms, 2) the data-driven component (\(DD\)) that represents the tentative norms. This separation allows us to inspect the mandatory norms and force the RL agent to take complying actions. Although there exist hybrid approaches that allow specifying norms manually as well as learning them \cite{osti_10301363,Baert2023} these approaches require the norms to be expressible in the same framework. However, this can be disadvantageous \cite{Rossi_Mattei_2019}. For example, \cite{10.1007/978-3-031-21203-1_5} discusses how LTL logic, often used in safe RL, may not be able to capture certain normative systems. HAVA's components are independent of each other thus allowing them to be specified using different techniques. Finally, approaches that specify all the norms manually \cite{Neufeld2022} will likely struggle to capture all necessary social norms due to their volume and evolving nature. For this reason we equipped HAVA with the data-driven component to enable learning the tentative norms.

\textbf{Properties of our agent:} \citet{Kasenberg_Scheutz_2018} propose that norm representations within the agent should be 1) context-dependent, 2) communicable, 3) learnable. The first requirement is fully satisfied by HAVA as it relies on a context as its input, here the state space of the given MDP.

The second requirement is satisfied by the \(RB\) component of the HAVA's Alignment Value \(\mathcal{AV}\). We believe that as \(RB\) describes the safety and legally critical norms there is a higher desire for \(RB\) to be communicable than in the case of the social norms in \(DD\). For this reason we have defined \(RB\) as a rule-based system. In HAVA, \(DD\) is not considered communicable as it can take many forms including neural networks, as per the experiments performed in this paper, which are notoriously hard to explain.

The third requirement is satisfied by the \(DD\) component. The social norms are learnt due to their high number, diversity and changeability. We expect the safety / legal norms in \(RB\) to be specified manually.

\textbf{How difficult is it to find \(\tau\) and \(\alpha\)?} While finding the right values of \(\tau\) and \(\alpha\) will likely require some experimentation, we believe we have identified several pointers to make it easier. Similarly to \citet{10.1145/3278721.3278728}, we argue for an evaluation that holds autonomous agents to the same standards as a human agent. This can then directly assist us in choosing the \(\tau\) hyperparameter; by asking ``what divergence to norms would still be indistinguishable from a human doing this task?'' we can arrive at a value of \(\tau\) as we have done in our experiment.

As for \(\alpha\) we can think of it in terms of the number of steps it takes to recover the reputation back to 1. We showed how this can be helpful for both environments with sparse and dense rewards. For sparse rewards (such as our toy grid world) \(\alpha\) should be low enough to cover the length of the value aligned trajectories. For dense rewards (such as our junction scenario) \(\alpha\) should be low enough to cover a significant portion of the produced trajectories. For example, settting \(\alpha=0.1\) (45 steps) covers about 20\% of the 220 actions the policy takes in the junction environment and it was sufficient, thanks to the dense rewards, to produce a value aligned policy.


\section{Limitations and Future Work}

One limitation of HAVA is its inability to distinguish antisocial but legal / safe behaviours from social legal / safe behaviours. Any behaviour in \(DD\) that is compliant with \(RB\) is discoverable by the HAVA agent. This could be potentially overcome by methods that take into consideration the frequency of observed behaviours such as \cite{10.5555/3504035.3504241} and could be an interesting direction for future work.

Automatic tuning methods for finding the \(\tau, \alpha\) hyperparameter values could be another direction for future work. This would eliminate the need for the human designer to set them making HAVA fully automatic.


\section{Conclusion}

In this paper, we presented the Hybrid Approach to Value Alignment (HAVA). This method defines an Alignment Value \(\mathcal{AV}\) which allows mixing of rule-based safety / legal norms with the learned social norms. This proposed Alignment Value, relies on two separate sets of norms: rule-based norms \(RB\) and learned norms \(DD\). The rule-based norms are expected to contain mostly safety-critical and legal norms that the agent should not violate while the learned norms are expected to approximate what a socially acceptable behaviour looks like. As the agent explores the environment, its reputation changes based on how much it violates the norms.

In a series of experiments, we have demonstrated the importance of HAVA's Alignment Value \(\mathcal{AV}\)'s reliance on both rule-based and data-driven norms. On one hand, rule-based approaches cannot hope to encode all the existing norms (such as the social norms). In our junction experiment we demonstrated that the absence of social norms can lead to socially unaccaptable, albeit safe / legal, behaviours. On the other hand, the data-driven approaches struggle to understand the severity of the learned norms and cannot distinguish those that should not be violated from those that on occasion can. To demonstrate this, we gave the data-driven agent a chance to violate its norms and observed that safety norms were violated in the process. Our proposed method solves both of these issues and produces policies that do not violate the rule-based norms while respecting the data-driven norms enough to pass for a human.

\section*{Acknowledgements}
This work was supported by UK Research and Innovation MINDS Centre for Doctoral Training [EP/S024298/1]. The authors acknowledge the use of the IRIDIS High Performance Computing Facility, and associated support services at the University of Southampton, in the completion of this work.

\bibliographystyle{ACM-Reference-Format} 
\bibliography{bibliography}


\end{document}